\documentclass[10pt,journal,compsoc]{IEEEtran}
\ifCLASSOPTIONcompsoc
  \usepackage[nocompress]{cite}
\else
  \usepackage{cite}
\fi
\ifCLASSINFOpdf
\else
\fi
\usepackage{graphicx}                
\usepackage{booktabs}                  
\usepackage{array,multirow}

\usepackage{amsmath}
\usepackage{url}

\begin{document}
%
\title{Enhanced Self-Perception in Mixed Reality: \\ Egocentric Arm Segmentation and Database with Automatic Labelling}

\author{Ester~Gonzalez-Sosa,
        Pablo~Perez,
        Ruben~Tolosana,
        Redouane~Kachach
        and~Alvaro~Villegas,
\IEEEcompsocitemizethanks{\IEEEcompsocthanksitem E. Gonzalez-Sosa, P. Perez, R. Kachach, and A. Villegas are Nokia Bell Labs Spain,
Maria Tubau 9, Madrid, 28050. 
E-mail: ester.gonzalez@nokia-bell-labs.com. 
\IEEEcompsocthanksitem R. Tolosana is with Universidad Autonoma de Madrid, E-mail: ruben.vera@uam.es}
\thanks{Manuscript received January 30, 2020; revised 2020}}
\markboth{Journal of \LaTeX\ Class Files,~Vol.~14, No.~8, August~2015}%
{Shell \MakeLowercase{\textit{et al.}}: Bare Demo of IEEEtran.cls for Computer Society Journals}
\IEEEtitleabstractindextext{
\begin{abstract}\rightskip=0pt In this study, we focus on the egocentric segmentation of arms to improve self-perception in Augmented Virtuality (AV). The main contributions of this work are: $i)$ a comprehensive survey of segmentation algorithms for AV; $ii)$ an \textit{Egocentric Arm Segmentation Dataset}, composed of more than $10,000$ images, comprising variations of skin color, and gender, among others. We provide all details required for the automated generation of groundtruth and semi-synthetic images; $iii)$ the use of deep learning for the first time for segmenting arms in AV; $iv)$ to showcase the usefulness of this database, we report results on different real egocentric hand datasets, including GTEA Gaze+, EDSH, EgoHands, Ego Youtube Hands, THU-Read, TEgO, FPAB, and Ego Gesture, which allow for direct comparisons with existing approaches utilizing color or depth. Results confirm the suitability of the EgoArm dataset for this task, achieving improvement up to $40\%$ with respect to the original network, depending on the particular dataset. Results also suggest that, while approaches based on color or depth can work in controlled conditions (lack of occlusion, uniform lighting, only objects of interest in the near range, controlled background, etc.), egocentric segmentation based on deep learning is more robust in real AV applications.
\end{abstract}
}

\maketitle

\IEEEdisplaynontitleabstractindextext

%
\IEEEpeerreviewmaketitle

\IEEEraisesectionheading{\section{Introduction}\label{sec:introduction}}

%
%
%
%

\IEEEPARstart{M}{ost} computer vision applications are traditionally focused on second or third point-of-view (POV),~actions that happen while interacting directly or indirectly with a camera, respectively \cite{bandini2019analysis}. With the advent of new wearable devices such as GoPro, Microsoft SenseCam, or even some Head Mounted Displays (HMD) useful for immersive applications, research on first-person POV or egocentric vision attracts some attention \cite{betancourt2015evolution}. Main research lines in egocentric vision can be categorized into:



\begin{itemize}
\item \textbf{Localize egocentric objects} usually involving knowing hand position and recognizing which objects are in contact with them. Typical tasks here are recognition \cite{ren2009egocentric}, detection \cite{li2013model}, segmentation \cite{serra2013hand}, tracking, and prediction \cite{8794474}, etc.
\item \textbf{Recognize the activities} performed by humans by analyzing the relationship between objects and hands (hand interaction) \cite{fathi2011understanding,cornacchia2016survey} or recognizing hand gestures for VR/AR \cite{rautaray2015vision,cheng2015survey,Tejo_2018}, hand poses \cite{li2019survey}.
\item \textbf{Visual lifelogging,} which consists of capturing daily live experiences \cite{bolanos2016toward}. Video summarization of people lives is also a related area, which could be used for detecting novel or anomalous events. This line is of special relevance for people with memory loss problems \cite{del2016summarization}.
\end{itemize}

\begin{figure*}[t]
  \centering
  \includegraphics[width=\linewidth]{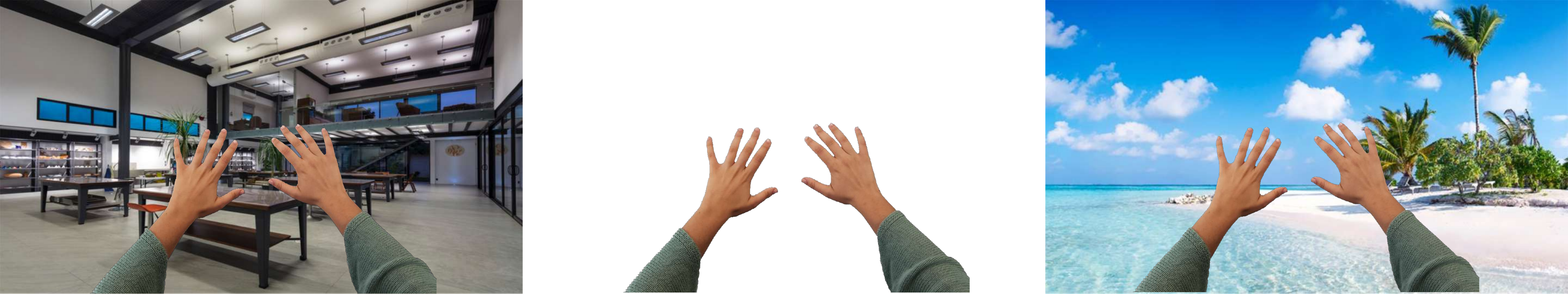}
  \caption{We propose semantic segmentation networks to segment human body parts (in this study whole arms) to get an enhanced self-perception in AV. Left: local reality; center: segmented arms; and right: AV with egocentric arms.}
	\label{fig:teaser}
\end{figure*}

In this study, we explore egocentric arm segmentation as an essential requirement for enhanced self-perception in Mixed Reality (MR)  (see Fig.~\ref{fig:teaser}). One of the main problems of immersive environments (IE)\footnote{Immersive Environment covers Virtual Reality (computer generated or 360$^{\circ}$ video), and also Mixed Reality, combining IE with the reality surrounding the user (local reality).} is the so-called presence factor: the subjective experience of being in one remote place without moving from the physical place. According to Lee \cite{lee2004presence}, the presence concept can be spread into three components: physical, social, and self-presence. In particular, self-presence involves \textit{experiencing the representation of one's own genuine self, physically or psychologically manifested, inside a virtual environment}.

First attempts towards self-perception on IE are based on avatars, which are virtual representations of the user mimicking his movements \cite{renner2013perception}. Current research line studies best avatar representations \cite{waltemate2018impact,argelaguet2016role}, their effect on the user \cite{aymerich2014relationship,schwind2017these} and when interacting with the IE \cite{aymerich2014relationship,khan2017evaluating,ries2008effect,feuchtner2017extending,gonzalez2019individual}.


Considering MR in particular, there is a different way of reaching self-perception. As stated by Milgram and Kishino \cite{milgram1994taxonomy,regenbrecht2004using}, AV is a MR subcategory of the virtuality continuum that aims to merge the reality surrounding the user (hereinafter local reality) with an IE. 
This means, instead of seeing an avatar of the user's body tracking his movements, the user is presented with his real body immersed in the IE. 
The merge of a real and virtual world can be achieved with the video see-through capabilities of the newest HMD devices such as HTC VIVE Pro or just by attaching a local camera to the HMD.
Hence, human body parts (such as hands, arms, lower body, etc.) or local objects (such as keyboards, smartphones, coffee cups, etc.) can be segmented from the see-through video and merged into the IE.
According to the objects being segmented, AV could be used to: $i)$ integrate self-presence and/or awareness of other people to prevent isolation, $ii)$ ease interaction with local objects \cite{mcgill2015dose}, or $iii)$ both.


Main segmentation approaches proposed in the literature for AV have been based on color or depth. However, they show still some limitations such as very complex physical setups \cite{mcgill2015dose}, limited field of view \cite{lee2016enhancing} or poor depth estimation that prevent AV from reaching its full potential. To overcome these limitations, we explore Semantic Segmentation algorithms (hereinafter Sem-Seg) proposed in the literature based on deep learning to segment egocentric \textbf{arms}. Our main motivation for focusing on whole arms and not just hands is an attempt to study this problem considering real life conditions. Indeed, arms and not just hands are easily visible when wearing a HMD. Besides, we hypothesize also that seeing your whole arms and not just your hands, may have a positive impact on the self-presence factor of the experience. Aside from the segmentation challenges pertaining to egocentric vision, the reader should notice that arms contain additional variability factors such as clothes or skin color that need to be considered. The proposed work is a continuation of some previous work \cite{gonzalez2018towards} that we significantly extend by the following contributions:



\begin{itemize}
\item a \textbf{comprehensive discussion} on segmentation methods for AV, categorized by color-based, depth-based and other approaches.
\item  an \textbf{Egocentric Arm Segmentation Dataset}, composed of more than $10,000$ semi-synthetic images, which is available for research purposes\footnote{https://cloud.proinnovation.es/index.php/s/tekqtneGXgrUgFD}. We describe the procedure carried out to automatically generate the groundtruth mask. 
\item proposal of \textbf{deep segmentation networks} designed to segment egocentric arms. To the best of our knowledge we are the first to consider them for AV applications and the first to consider whole arms and not just hands.
\item a \textbf{thorough evaluation} of our method to a wide number of real egocentric datasets existing in the literature: GTEA Gaze+ \cite{li2015delving}, EDSH \cite{Li_2013_CVPR}, EgoHands \cite{Bambach_2015_ICCV}, EgoYoutubeHands \cite{urooj2018analysis}, THU-READ \cite{tang2018multi}, TEgO \cite{lee2019hands},  FPAB \cite{garcia2018first}, and EgoGesture \cite{zhang2018egogesture}. We contribute also with a segmentation groundtruth of a representative subset of $277$ images from EgoGesture, which will also be made available for research purposes.
\item a \textbf{comparison with former approaches} based on color or depth, highlighting their pros and cons. 
\end{itemize}


\begin{table*}[h]
\tiny
\caption{Related works on Egocentric Human Body Parts or Object Segmentation for Augmented Virtuality. They are categorized into color-based, depth-based or other approaches. CRF stands for Conditional Random Fields. Notice that \cite{serra2013hand,urooj2018analysis} are focused on egocentric hand segmentation but are not designed for AV.}
  \label{tab:related_av}
  \begin{tabular}{cccccccc}
    \toprule
    Work & Category & Framework & Augmented Element & Segmentation Method & Observations & Survey\\
    \midrule
    

     \begin{tabular}[c]{@{}c@{}} Metzger \textit{et al.} $1993$ \\ \cite{metzger1993adding}\end{tabular}  & \multirow{10}{*}{Color} &  \begin{tabular}[c]{@{}c@{}} Virtual Research Flight Helmet \\ Toshiba IK-M40A cameras \end{tabular} &  Hands &  Blue chroma-key & \begin{tabular}[c]{@{}c@{}} Illumination \\ dependent \end{tabular} & No \\[0.4cm] 
     
          \begin{tabular}[c]{@{}c@{}} McGill \textit{et al.} $1993$ \\  \cite{mcgill2015dose}\end{tabular}  &  & \begin{tabular}[c]{@{}c@{}}  Oculus Rift DK1 \\ 1.8mm M12 wide-angle board lens \end{tabular} &  \begin{tabular}[c]{@{}c@{}}  Hands  \\ Keyboard \end{tabular} &  Green chroma-key  & \begin{tabular}[c]{@{}c@{}} Illumination \\ dependent \end{tabular} & Yes ($108$ subjects) \\[0.4cm]

      \begin{tabular}[c]{@{}c@{}} Bruder \textit{et al.} $2009$ \\ \cite{bruder2009enhancing}\end{tabular} & &  \begin{tabular}[c]{@{}c@{}}eMagin 3DVisor Z800 \\ 2 USB cameras\end{tabular} &  \begin{tabular}[c]{@{}c@{}}Hands \\ Body \end{tabular} &  \begin{tabular}[c]{@{}c@{}} Skin detection \\ Floor subtraction \end{tabular} & \begin{tabular}[c]{@{}c@{}} Brightness \\ adjustment \end{tabular} & Yes ($7$ subjects)\\[0.4cm]

              \begin{tabular}[c]{@{}c@{}} Perez \textit{et al.} $2018$ \\ \cite{immersirve_gastronomic2019}\end{tabular} & & \begin{tabular}[c]{@{}c@{}} Samsung Gear HMD \\ Samsung S8  \end{tabular} &  \begin{tabular}[c]{@{}c@{}} Hands \\ and food \end{tabular} &  \begin{tabular}[c]{@{}c@{}} Color based on Cr from YCrCb  \end{tabular} & \begin{tabular}[c]{@{}c@{}} Scalable \\ to other objects \end{tabular} & Yes  ($66$ subjects)\\[0.4cm] 
            
            \hline
            
              \begin{tabular}[c]{@{}c@{}} Tecchia \textit{et al.} $2014$ \\ \cite{tecchia2014m}\end{tabular}  & \multirow{16}{*}{Depth} & \begin{tabular}[c]{@{}c@{}} Oculus DK1 \\ 3D camera \end{tabular} &  \begin{tabular}[c]{@{}c@{}}Hands \\ Body \end{tabular} &  Depth & \begin{tabular}[c]{@{}c@{}} Virtual objects manipulation \\ through fingertip tracking \end{tabular} & No\\[0.4cm] 
            
              \begin{tabular}[c]{@{}c@{}} Nahon \textit{et al.} $2015$ \\ \cite{nahon2015never}\end{tabular} &  & \begin{tabular}[c]{@{}c@{}} Oculus Rift DK1 \\ Kinect v2 \end{tabular} &  \begin{tabular}[c]{@{}c@{}} Own Body \\  Local objects \\ Other People \end{tabular} &  Depth & \begin{tabular}[c]{@{}c@{}} Kinect placed opposite \\ to the user \\  Point cloud \\ not very dense \end{tabular} & No\\[0.4cm]

             \begin{tabular}[c]{@{}c@{}} Lee \textit{et al.} $2016$ \\ \cite{lee2016enhancing}\end{tabular}  & &  \begin{tabular}[c]{@{}c@{}} Oculus Rift DK2 \\ Soft Kinect DS325 \end{tabular} &  User's body &  Depth & \begin{tabular}[c]{@{}c@{}} Head shaking \\ for transition \end{tabular} & Yes  ($14$ subjects)\\[0.4cm] 
             
                 \begin{tabular}[c]{@{}c@{}} Alaee \textit{et al.} $2018$ \\ \cite{alaeeuser}\end{tabular} & &  \begin{tabular}[c]{@{}c@{}} Oculus Rift DK2 \\ Intel Real Sense \end{tabular} &  \begin{tabular}[c]{@{}c@{}} Hands \\ and smartphone \end{tabular} &  \begin{tabular}[c]{@{}c@{}} Depth in $10-40$ cm \\  \end{tabular} & \begin{tabular}[c]{@{}c@{}} Scalable \\ to other objects \end{tabular} & Yes  ($25$ subjects)\\[0.4cm]

                 \begin{tabular}[c]{@{}c@{}} Rauter \textit{et al.} $2019$ \\ \cite{rauter2019augmenting}\end{tabular} & & \begin{tabular}[c]{@{}c@{}} HTC Vive Pro \end{tabular} &  \begin{tabular}[c]{@{}c@{}} Near Real \\ World Objects \end{tabular} &   Depth + post processin2g  & \begin{tabular}[c]{@{}c@{}} Flexible \\ depth range \end{tabular} & No \\[0.4cm] 
                                   
           \hline
      
                \begin{tabular}[c]{@{}c@{}} Serra \textit{et al.} $2013$ \\ \cite{serra2013hand}\end{tabular} & \multirow{15}{*}{Other} &  --&  Hands &  \begin{tabular}[c]{@{}c@{}} Random Forest \\ Light + Time + Space Consistency  \end{tabular} & \begin{tabular}[c]{@{}c@{}} Gesture Recognition \end{tabular} & No\\[0.4cm]     

           \begin{tabular}[c]{@{}c@{}} Desai \textit{et al.} $2017$ \\ \cite{desai2017window}\end{tabular} &  &  \begin{tabular}[c]{@{}c@{}} Oculus Rift DK2 \\ Leap Motion \end{tabular} &  Smartphone &  \begin{tabular}[c]{@{}c@{}} Smartphone edge detection +\\ statistical classifier + \\ App sending screenshots  \end{tabular} & \begin{tabular}[c]{@{}c@{}} No scalable \\ to other objects \end{tabular} & No\\[0.4cm]      
         
            \begin{tabular}[c]{@{}c@{}} Korsgaard \textit{et al.} $2017$ \\ \cite{korsgaard2017immersive}\end{tabular} & & \begin{tabular}[c]{@{}c@{}} Oculus Rift CV1 + Touch \\ OVRvision PRO\end{tabular} &  \begin{tabular}[c]{@{}c@{}} Local reality \\ with food  \end{tabular} &  \begin{tabular}[c]{@{}c@{}} Head orientation:  \\ angle in 25-30 $\deg$ \end{tabular} & \begin{tabular}[c]{@{}c@{}} No optimal \\ user immersion \end{tabular} & Yes  ($6$ subjects)\\[0.4cm] 
            
            \begin{tabular}[c]{@{}c@{}} Urooj \textit{et al.} $2018$ \\ \cite{urooj2018analysis}\end{tabular} &  & -- &  Hands in the Wild &  \begin{tabular}[c]{@{}c@{}} DeepLearnig\\ (RefineNet +CRF)  \end{tabular} & \begin{tabular}[c]{@{}c@{}} Cross dataset \\ evaluation \end{tabular} & No\\[0.4cm]    
                       
                             
		Proposed Approach  &  & \begin{tabular}[c]{@{}c@{}} Samsung Gear HMD \\ Samsung S8 \end{tabular} &  Arms &  \begin{tabular}[c]{@{}c@{}} Deep Learning \\  (DeepLabv3+)  \end{tabular} & \begin{tabular}[c]{@{}c@{}} Re-training \\ for Scalability \end{tabular} & No\\[0.4cm] 
      
  \bottomrule
\end{tabular}
\end{table*}

The rest of this article is structured as follows. Section~\ref{related_works} covers related works regarding AV, with an emphasis on the different algorithms that have been proposed to date to segment local reality objects. Section~\ref{VR_hand_dataset} describes the \textit{Egocentric Arm Segmentation Dataset} and the whole procedure to generate semi-synthetic images while automatically obtaining the segmentation groundtruth. Section~\ref{semantic_segmentation} presents the Sem-Seg algorithms considered to segment egocentric arms. Then, Section~\ref{ep} explains the experimental protocol and test datasets considered to conduct the experiments, while Section~\ref{results} report the segmentation results and the comparison with former segmentation approaches used for AV. Finally, Section~\ref{conclusions} concludes the paper with some discussions.

\section{Previous Segmentation Approaches for Augmented Virtuality}
\label{related_works}

 \begin{figure*}[t]
  \centering
  \includegraphics[width=0.8\linewidth]{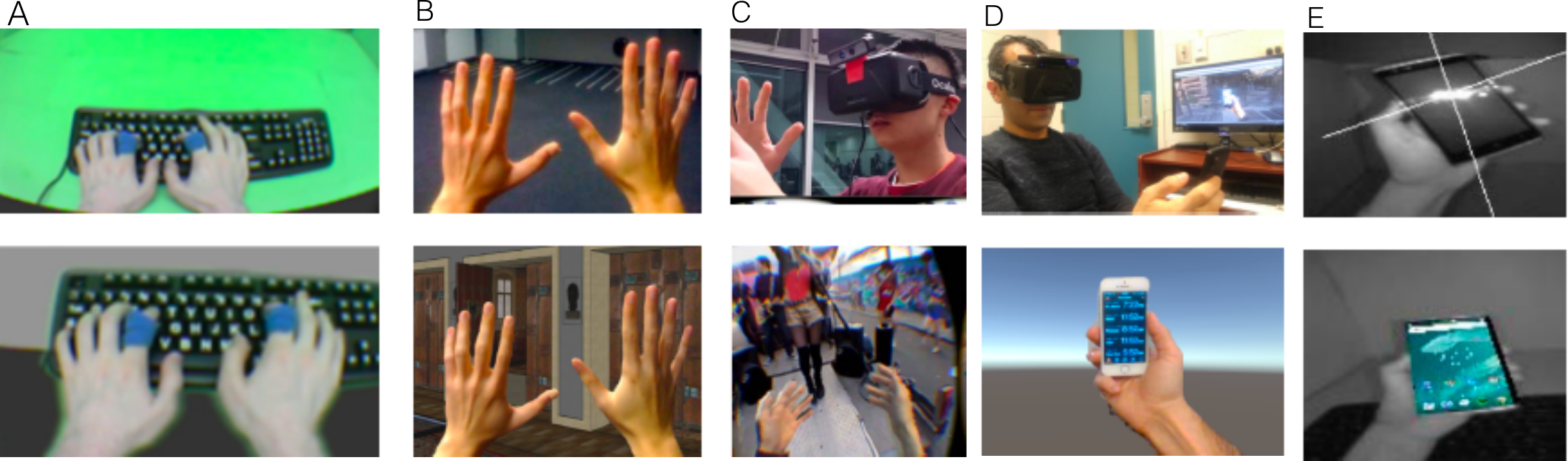}
  %
  %
  \caption{\label{fig:av_related_works}
           Examples of the different segmentation approaches proposed in the literature to segment particular objects from the local reality to be blended with VR. From left to right: $i)$ green chroma-key \cite{mcgill2015dose}, $ii)$ skin detection \cite{bruder2009enhancing}, $iii)$ and $iv)$ depth information \cite{lee2016enhancing,alaeeuser} and $v)$ edge detection and statistical classifier \cite{desai2017window}}
\end{figure*}

Our aim within this section is to undertake a comprehensive and thorough review of related works regarding how to segment local reality objects. Table \ref{tab:related_av} lists the most relevant works published in the scientific literature in this regard (for information regarding Sem-Seg from the state-of-the-art, we refer to \cite{garcia2017review}). We proceed now to describe them based on the segmentation method used.


\subsection{Color-based}

One of the preliminary approaches for segmenting objects from local reality was the chroma-key, similar to the concept applied within weather forecast in television for decades. The idea is simple: given an input video with this chroma-key color presented, only pixels not sharing this color are retained. 
Metzger \textit{et al.} \cite{metzger1993adding}, one of the pioneers of the idea of AV, put forward the use of blue chroma-key, to select the user's hands from the local reality. Further, the authors pointed out the importance of having the space uniformly lit to obtain accurate results. Similarly, McGill \textit{et al.} \cite{mcgill2015dose} used a green chroma-key to filter objects from the local reality (see Fig.\ref{fig:av_related_works} A). The particular task involved typing a keyboard in a VR environment. For this purpose, they designed a scenario with a green chroma surface where the keyboard was placed. The segmentation was performed in two stages: first, both hands and the keyboard were segmented by discarding all pixels that share the green color; then hand detection was carried out using blob detection and the help of some hand markers to recognize keyboard actions. Although results obtained with this simple method were almost perfect in terms of segmentation, the application itself is very limited if the local reality appearance is constrained to exhibit a certain chroma-key color.  

Focusing particularly on the hand segmentation problem, researchers also proposed the use of skin detection algorithms to segment hands from local reality \cite{bruder2009enhancing} (see Fig.\ref{fig:av_related_works} B).~The idea behind this algorithm is the following: the local reality image is first transformed to the HSV color space, and then it is filtered out so that only pixels values that are around a certain Hue range ($\mu \pm \sigma$) are segmented. Although this approach enhanced the green chroma-key approach in the sense that local reality is not constrained anymore, some false positives may appear having similar skin color such as it is the case of faces in the scene, furniture, boxes, etc. In the same work, lower body part that could be seen from the egocentric view was also segmented with a naive floor subtraction approach. Taking the assumption that the floor appearance was uniform, the body was retained by simply filtering out all pixels not belonging to the floor.

Likewise,~Perez \textit{et al.} \cite{immersirve_gastronomic2019} used a YCbCr skin detection algorithm based on red chrominance, adding a transparency alpha layer to the local reality. By using less strict thresholds than those normally used for skin detection, objects with yellow and red tones with high saturation such as food were also segmented. This segmentation method allowed them to build a proof of concept of an Immersive Gastronomic Experience using Distributed Reality \cite{DR2019}, a new type of Mixed Reality that involves capturing different realities (~at least one remote in the form of $360^{\circ}$ video and a local reality) to foster remote human communication and shared experiences.

There are some inherent limitations in color-based approaches: they require specific physical setups, where no background objects have any of the colors included in the foreground (this is especially restrictive in traditional green chroma), and they are very sensitive to illumination conditions.



\subsection{Depth-based}

Based on the idea of filtering out everything that is below a certain depth threshold value (segmentation of a user-centered bubble), Nahon \textit{et al.} \cite{nahon2015never} blended into the IE not only the user's own body but also objects from the local reality and even other people. This way, self-presence is increased and also interaction and communication with other objects or people is feasible. Likewise, Lee \textit{et al.} \cite{lee2016enhancing} used depth information to include the user's own body into an immersed cinema experience (see Fig.\ref{fig:av_related_works} C). They also incorporated interactivity so that the level of user embodiment was adapted to the content or to user preferences. Alaee \textit{et al.} \cite{alaeeuser} also incorporated objects which were in the distance range of $10-40$ cm, with the aim of interacting with the smartphone in the IE (see Fig.\ref{fig:av_related_works} D). More recently, Rauter \textit{et al.} \cite{rauter2019augmenting} implemented the same idea while estimating depth from the stereo camera of HTC VIVE Pro.~They also performed some post-processing of the estimated foreground mask to address pixels with missing depth values. 

Depth-based solutions are relatively simple to implement, due to the affordability of RGB-D sensors. However, such sensors have some limitations: on the one hand, depth estimation is noisy and prone to artifacts when handling near objects, specular materials, non-reachable areas, or shadows \cite{moya2017modeling}; on the other, they have a very narrow field of view which also impairs sense of presence \cite{lee2016enhancing}. 



\subsection{Other approaches}

Aside from the mainstream segmentation approaches, alternative ones have been proposed. For instance, Desai \textit{et al.} \cite{desai2017window} proposed a method to segment smartphones or tablets based on two stages: $1)$ edge-based object detection to select the smartphone; and $2)$ a statistical classifier based on attributed features decided whether the segmented object was a smartphone (see (see Fig.\ref{fig:av_related_works} E). The overall aim was to allow interactivity with these devices while being immersed. This algorithm, however, is not scalable to segment other objects. 

Korsgaard \textit{et al.} \cite{korsgaard2017immersive} conducted an AV experience in which the user had to interact with real food placed in front of him. Merge between both worlds was achieved through head orientation. Everytime the head was orientated in a downward angle (where food is normally placed), the local reality was visible, whereas if the user looked straight ahead, the IE became visible. The main limitation of this approach is that no optimal full immersion is achieved but just an angle-based transition approach between the IE and the local reality.

Beyond using skin color, there are other attempts, not specifically designed for AV, to segment hands from an image-based point of view. Serra \textit{et al.} \cite{serra2013hand} proposed a hand-crafted method for segmenting skin color based on random forest superpixel classification considering, light, time and space consistency. Although it may be seen as an evolution of color-based methods, this approach still would fail to segment arms containing clothes. There are also some attempts to detect \cite{Bambach_2015_ICCV} or segment hands \cite{urooj2018analysis} using deep learning that shows the feasibility of adapting existing pre-trained models such as RefineNet or CaffeNet (slight modified version of Alexnet). Again, these approaches are focused on segmenting hands, and not whole arms.

\section{EgoArm: Egocentric Arm Segmentation Dataset}
\label{VR_hand_dataset}

\begin{table}[b]
\scriptsize
  \caption{Egocentric Arm Segmentation Dataset.}
  \label{VR_Hand_Dataset}
  \begin{tabular}{cc}
    \toprule
    Feature & Values\\
    \midrule
    People  & $9$ male and $4$ female                   \\ 
    Arm pose & close hands, open palm, open dorsum, left arm, right arm  \\ 
    Scenario      & outdoors, indoors                                        \\ 
    Outfit        & outfit1, outfit2                                         \\ 
   Sleeve        & Long sleeve, short sleeve                                  \\     
     Ethnicity             & Caucasian, Black, Mixed      \\
  \bottomrule
\end{tabular}
\end{table}



At the present time of writing, there are not databases in the literature suitable for egocentric arm segmentation. There exist some databases that are related to egocentric hand detection or segmentation but not related to the whole arm. Therefore, we introduce the \textit{Egocentric Arm Segmentation Dataset (EgoArm)}, which is designed with a wide range of variations to maximize generalization capabilities. Table \ref{VR_Hand_Dataset} describes the main characteristics of EgoArm, containing more than $10,000$ images. We highlight that EgoArm includes images of people with different skin color and gender.

 \begin{figure*}[t]
  \centering
  \includegraphics[width=.9\linewidth]{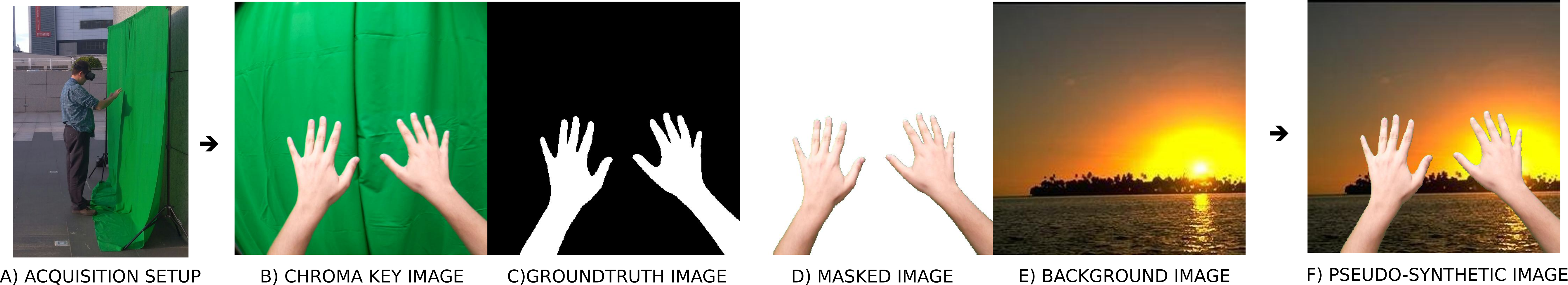}
  %
  %
  \caption{\label{fig:database_creation}
           Procedure to obtain groundtruth and semi-synthetic images: through an Android app installed in the smartphone, images are recorded from the HMD perspective using a chroma-key approach. Subsequently, we applied HSV filtering to obtain the groundtruth images. With the groundtruth image, we select the relevant information from the chroma-key image that will be later combined with a background image to form the final semi-synthetic image.}
\end{figure*}

 \begin{figure*}[t]
  \centering
  \includegraphics[width=.9\linewidth]{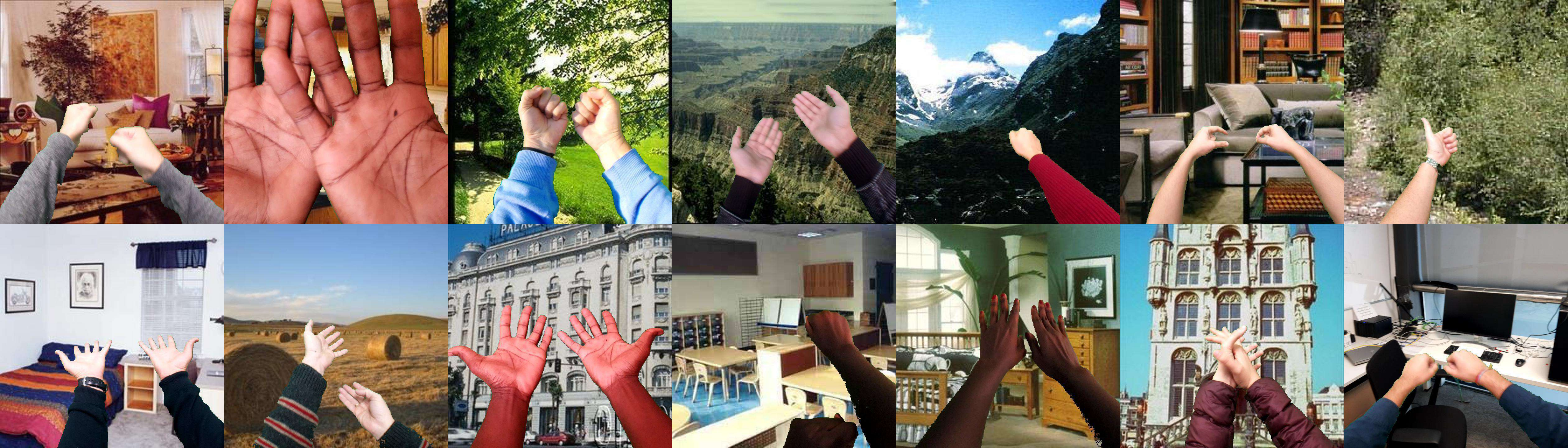}
  %
  %
  \caption{\label{fig:VR_hand_samples}
           Example Images of the Egocentric Arm Segmentation Datase showing a wide variety in terms of different subjects, gender, arms position, scale, clothes, skin color, illumination and background.}
\end{figure*}
%

Unlike other supervised learning approaches such as classification or regression, in which required labels or groundtruth are just text labels or a few numbers defining bounding boxes, Sem-Seg labels are images where every pixel contains a particular number accounting for the class information. The acquisition of such databases is time-consuming, which represents a major problem that has already been observed by Bandini and Zariffa \cite{bandini2019analysis}. To overcome this issue, we propose a semi-automatic way of labelling images (see Fig.\ref{fig:database_creation}), composed of the following steps:








\subsection{Acquisition} An Android application is developed in order to record $30$ fps videos from the Samsung-S8 frontal camera while the subject is wearing the Gear VR Samsung headset with the smartphone in front of a chroma-key backdrop (see Fig.\ref{fig:database_creation}A). Unlike other segmentation datasets, we decided to record videos at $720\times720$ in order to target the high resolution requirements of VR applications (Fig.\ref{fig:database_creation}B). Each session is designed to record videos with a particular configuration in terms of people, scenario, outfit, and sleeve. A recorded assistant ensures that, at each session, videos from the five different arm poses are recorded.  

%

\subsection{HSV Filtering} With the recorded chroma-key videos (see Fig.\ref{fig:database_creation}B.), a HSV-based filter is applied to obtain the foreground images (values are in the range $0-1$), as follows:

\begin{equation} \label{green_chroma}
    f(x)= 
\begin{cases}
    1 & \text{if } H(x,y)\leq h_{1} \land H(x,y)\geq h_{2} \land  S(x,y)\geq s_{1} \\
    0              & \text{otherwise}
\end{cases}
\end{equation}

\noindent being $h_{1}$, $h_{2}$ and $s_{1}$ set to $0.22$, $0.45$, and $0.20$, respectively (values obtained by empirical testing). To prevent high similarity, images are selected every $5$ frames. Additionally, some morphological operations are applied to delete noisy areas (see Fig.\ref{fig:database_creation}C). 


\subsection{Masking} Before creating the semi-synthetic image, the chroma-key image is masked with the groundtruth image to get the area of interest (in this case: arms, see Fig.\ref{fig:database_creation}D.)

\subsection{Combination} Semi-synthetic images (Fig.\ref{fig:database_creation}F) are created combining background (Fig.\ref{fig:database_creation}E) with chroma-key images (Fig. \ref{fig:database_creation}B)) masked with foreground images (Fig.\ref{fig:database_creation}D). Natural background images are obtained from the MIT Scene Parsing Benchmark \cite{zhou2017scene}. Among the whole set of $20,210$ images, we select those which hold $height=width$ and then reshape it to $720\times720$, resulting in a subset of $3,697$ different background images. These backgrounds contain indoor scenes related to houses, public spaces, commercial places as well as outdoor scenes such as landscapes, beaches, mountain, etc. As a final post processing, we discarded those pair of \textit{groundtruth-semi-synthetic} images with some false positives in the groundtruth. Fig.\ref{fig:VR_hand_samples} shows examples of the variability of these images.





\section{Egocentric Arm Segmentation}
\label{semantic_segmentation}


Accurate and robust arm segmentation is vital to achieve enhanced self-perception in MR. DL-based approaches have been shown to outperform conventional approaches in diverse computer vision tasks if the used training data reflects real-world scenarios. This clearly motivates the development of a DL-based are segmentation system which is expected to outperform traditional approaches (see Section  \ref{related_works}) in terms of robustness. Convolutional Neural Networks (CNN) have been shown to be the state-of-the-art for supervised classification and detection tasks \cite{goodfellow2016deep}. CNNs are composed of different types of hidden layers: Convolutional, Rectifier Linear Unit, Pooling and Fully Connected (FC). FC are the final layers of CNNs that, along with the classification layer, hold the output (having the same size as the number of objects to classify). In $2015$, Long \textit{et al.}~proposed Fully Convolutional Networks (FCN): a modification of CNN architectures that reached state-of-the-art performance in Sem-Seg problems. Concretely, they replaced fully connected layers by fully convolutional ones to preserve the spatial dimension while keeping the class identity information \cite{long2015fully}. Another important key component aside from the encoding subnetwork here is the decoding subnetwork, which is placed after the fully convolutional layers and is entrusted of upsampling the class spatial map up to the original input size. 

\subsection{Considered Sem-Seg Networks}
\label{deep_network}


Our hypothesis, confirmed also by previous work \cite{rogez20143d}, is that segmentation networks trained for third POV fail when segmenting from egocentric vision. Indeed, egocentric vision has the advantage that the objects tend to appear at the center of the image, but also the challenge of the camera moving with the human body, which creates fast movements and sudden illumination changes.

Due to the relatively small size of EgoArm (in comparison with datasets aimed to train architectures from scratchs such as Imagenet, Pascal VOC, etc.), we took the decision to apply tranfer-learning from existing Sem-Seg architectures. The first Sem-Seg architecture considered was the FCN, proposed by Long \textit{et al.} and originally trained for the PASCAL VOC 2011 segmentation challenge, which consist of segmenting up to $20$ classes categorized into people, $6$ different animals, $7$ means of transport, and $6$ house objects. After having empirically found the best training parameters for fine-tuning the FCN architecture with the EgoArm database, we observed that the output mask was not giving accurate enough segmentation masks for the $720\times720$ required resolution.

The next architecture that we considered was DeepLab, originally proposed in $2017$ as a new architecture for Sem-Seg. In particular, among the different enhanced networks proposed since then \cite{chen2017rethinking,chen2018deeplab,chen2018encoder}, we consider here the DeepLabv3+ \cite{chen2018encoder} due to $i)$ the use of the Res-net pre-trained model, replacing the former VGG-16 pre-trained model, placing $4$ of these blocks in cascade. It is characterized by $101$ layers and the introduction of short-cut connections \cite{ronneberger2015u}; $ii)$ the use of \textit{a-trous} convolution, using upsampled filters that allow dense feature extraction taking context into account, without increasing the number of parameters; $iii)$ the use of Atrous Spatial Pyramid Pooling module to robustly segment objects at multiple scales; and $iv)$ the use of a decoder module to refine the segmentation results \cite{badrinarayanan2017segnet}, especially along object boundaries. In general, the fact that this arquitecture was very deep at the encoding subnetwork and deeper than the existing approaches in the decoding subnetwork give us the idea that could segment accurately  egocentric images of high resolution. In order to understand the gain achieved with the fine-tuned network with respect to the original DeepLabv3+ model, we decide to use two different semantic segmentation networks:

\begin{itemize} 
        \item \textbf{DeepLabv3+}: our idea here is to use the original DeepLabv3+ to segment egocentric arm and confirm our hypothesis.  This original network was trained using the PASCAL VOC database, so arms are segmented pertaining to people class.
        \item \textbf{DeepLabv3+ using EgoArm}: we apply transfer learning using images from the EgoArm dataset so that the network segments two classes: arms and background. As there are more male than female subjects and in order to have a more gender-balanced dataset, we discard $4$ male subjects, having a total of $11561$ images.
\end{itemize}

We also considered combining EgoArm images with a subset of images from PASCAL VOC containing people. Concretely, we select $4,344$ images among the entire PASCAL VOC dataset, resulting in a total of $15,905$ images. As we were only interested on the people class, we relabelled all remainder pixels associated to any of the other $20$ classes, to the background class. In this case, we have two classes: people and background, where arms are considered part of the people class. However, we did not find improvements for the egocentric segmentation tasks, as images from PASCAL VOC are acquired from a third-point-of-view perspective.

\section{Experimental Protocol}
\label{ep}



The GTEA Gaze+ dataset, which contains $1,115$ images, is used as the validation dataset \cite{Bambach_2015_ICCV}. The main motivation of not using a subset of EgoArm as the validation set, is that we aim to validate our results in a real egocentric database. Among the public real egocentric datasets, GTEA Gaze+ is the largest one and more similar to the arm segmentation task. It contains egocentric arms performing actions in a kitchen, with a very cluttered environment. In this dataset, groundtruth is related to the skin color but no clothes were presented in the images. 

When it comes to the training of deep neural networks using stochastic gradient descent algorithm, several hyperparameters need to be adjusted.  An exhaustive set of experiments following grid search strategies, have been conducted monitoring validation performance over the GTEA Gaze+. Training has been done using two GPU GTX-1080 Ti with $12GB$ RAM each. Batch size was set to $4$ due to the large size of the training images ($720\times720$). The final training of the DeepLabv3+ Ego Arm was achieved using an initial learning rate of $1e-3$, $2$ epochs, $7500$ as maximum number of iterations for reducing the learning rate, a final learning rate of $1e-6$, and weight decay of $1e-5$. 

\subsection{Tests}
\label{test_databases}

\begin{figure}[t]
  \centering
  \includegraphics[width=1.0\linewidth]{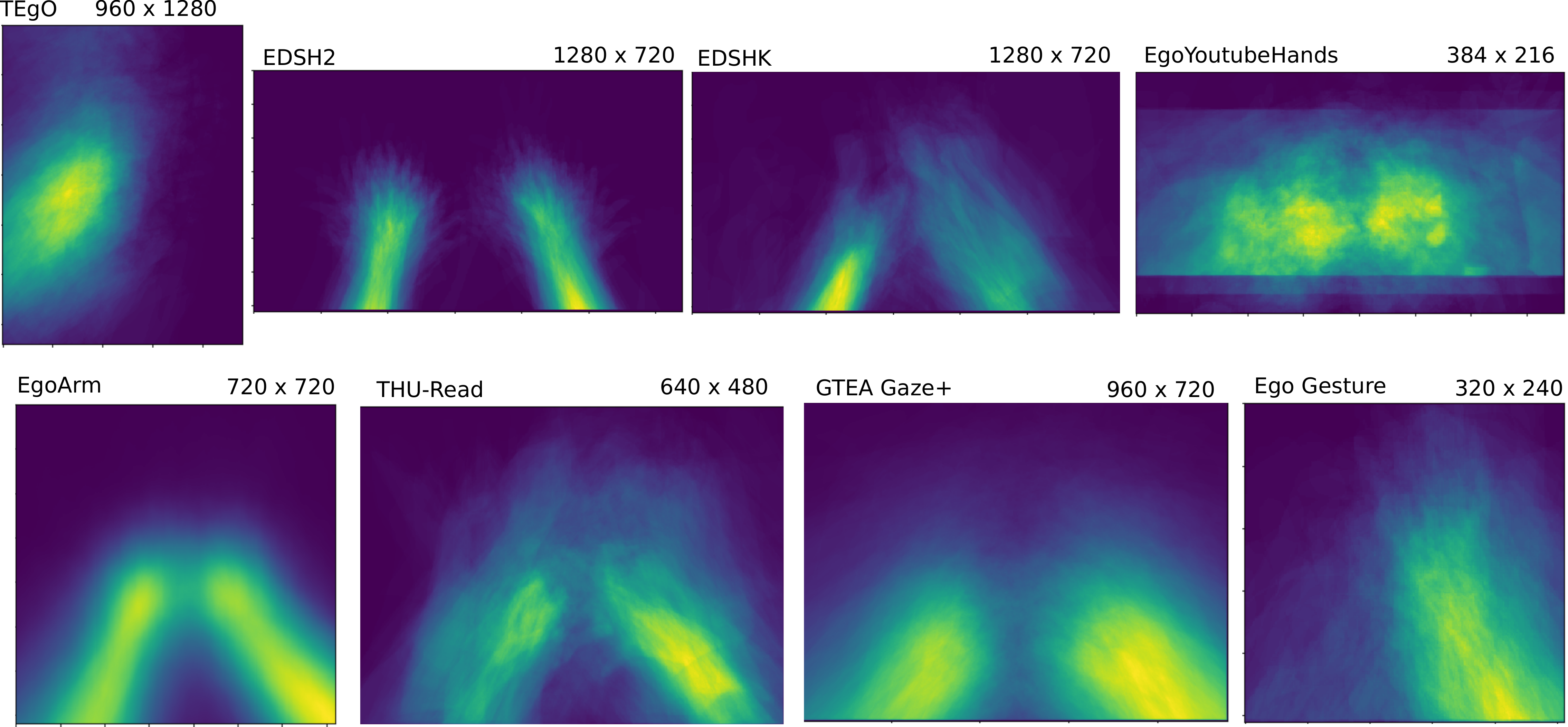}
  %
  %
  \caption{\label{fig:heatmaps}
    Heatmaps for Spatial Maps Occurencies for the different test datasets considered. From up to bottom and left to right: TeGO, EDSH2, EDSHK, EgoYoutubeHands, EgoArm, THU-Read, GTEA Gaze+, and Ego Gesture.}
\end{figure}

In order to assess the generalization capabilities of our algorithm, we perform the evaluation on different public datasets\footnote{There were also other datasets available in the literature that we discarded for different reasons. For instance, The EPIC-KITCHENS Dataset is not providing segmentation masks \cite{Damen_2018_ECCV}; the Egocentric Gesture Recognition dataset\cite{Tejo_2018} only provide segmentation masks for chroma-key hand gesture images; Keyboard Hand Dataset (KBH, \cite{wang2019recurrent}) was not found available for research use.}:


\begin{itemize}


\item \textbf{EDSH (groundtruth related to skin color)} \cite{Li_2013_CVPR}: EDSH2 and EDSH kitchen are the test videos of EDSH, and contain indoor and outdoor scenes with large variations of illumination, mild camera motion induced by walking and climbing stairs, with just $1$ user. They provide $104$ and $197$ segmentation masks for the test datasets EDSH2 and EDSHK, respectively.

\item \textbf{EgoHands, (groundtruth related to hands)} \cite{Bambach_2015_ICCV}: it contains $48$ Google Glass videos of complex, interactions between two people playing board games (one with first POV, and the other with thrid POV). In order to reduce redundancy and computational load, we create a subset of this dataset, by selecting $10$ images per each of the $48$ different videos, resulting in a total of $480$ images.

\item \textbf{Ego Youtube Hands (groundtruth related to hands)} \cite{li2015delving}: It contains $3$ egocentric videos from daily activities. Among the entire set of $1,032$ images, we create a subset including images showing hands and arms, resulting in a total of $689$ frames.


\item \textbf{TEgO database (groundtruth related to skin color)} \cite{lee2019hands}: in order to test the robustness against black skin color, we report results using the test set pertaining to subject $B1$ (which has black skin color), composed of different subsets of images under different illumination (normal and extreme) and background conditions (vanilla or in the wild). 

\item \textbf{THU-Read (groundtruth related to skin color)} \cite{tang2018multi}: initially created for egocentric action recognition from RGB-D data, they contain a subset of $650$ images with egocentric actions where users arms appears holding different objects. Images are of $640\times480$, but their original resolution is lower so images appear pixelated. 


\item \textbf{FPAB, First Person Action Bechmark (no groundtruth available)} \cite{garcia2018first}: dataset that provides both color and depth images from egocentric images. As their original purpose was to infer hand pose, people are wearing some reference marks on their hands. As the color and depth images were extracted from different sensors and at different positions, it was very difficult to create a common groundtruth, so we do not report empirical results, but only visual examples.

\item \textbf{Ego Gesture (groundtruth created in this work and related to whole arms)} \cite{zhang2018egogesture}: Ego Gesture database contain egocentric color and depth videos acquired from RealSense SR300. It contains $83$ different hand gestures from $50$ different subjects and $6$ different scenarios (e.g: indoors, outdoors, illumination, static clutter background, dynamic background, walking, etc.). We then create a subset of $277$ images, by selecting approximately one image per subject and different scenario. As the groundtruth of this dataset was related to the hand gesture, we manually labelled the segmentation mask of this subset of images, labeling the whole arm as groundtruth.

Notice that the groundtruth of the former datasets are related etiher to hands or skin color but not the arm concept itself. This discrepancy would be covered in Section \ref{results}. Fig. \ref{fig:heatmaps} describes the heatmaps of the different datasets; to give an idea of the type of groundtruth and the average position of hands/arms on those datasets.

\end{itemize} 

\begin{table*}[h]
\small
  \caption{Segmentation results in terms of Intersection over Union for different egocentric segmentation datasets. The segmentation algorithms considered are: $1)$ color-based, $2)$ original Deeplabv3+ using the person class; $3)$ Deeplabv3+ using the proposed Egocentric Arm Segmentation Dataset. GTEA Gaze+ is our validation dataset. The reader should bear in mind that there is discrepancy between the available groundtruth (related to hands or skin color) with the arm concept. Due to that Miss Rate is also reported. Bold indicates best IoU for a given dataset.}
  \label{tab:results}
  \begin{tabular}{ccccc}
    \toprule
    Database &  \# Images & Color  & DeepLabv3+ & DeepLabv3+ EgoArm  \\
    \midrule

    GTEA Gaze+ \cite{Bambach_2015_ICCV}  &  $1115$ &  $15.47$ ($4.13$) &  $40.57$ ($50.68$) & $\mathbf{60.75}$ ($7.50$)   \\[0.4cm] 
    
     EDSH2 +  \cite{Li_2013_CVPR}  &  $104$ & $52.62$ ($16.93$) & $67.56$ ($13.13$) & $\mathbf{74.04}$ ($8.30$)    \\[0.3cm] 
    
     EDSHK + \cite{Li_2013_CVPR}  &  $197$ & $42.24$ ($11.23$) & $52.50$ ($22.91$) & $\mathbf{56.61}$ ($8.10$)  \\[0.3cm] 
     
     
      EgoHands  \cite{Bambach_2015_ICCV}  &  $480$ &  $29.09$ ($16.19$) &  $26.65$ ($18.69$) & $\mathbf{33.52}$ ($17.85$) \\[0.3cm] 
    
         
            EgoYoutubeHands  \cite{li2015delving}  &  $689$ &  $\mathbf{22.11}$ ($34.35$) &  $20.64$ ($49.67$) & $20.80$ ($35.71$) \\[0.3cm] 
         
            
             Ego Gesture  \cite{zhang2018egogesture}   &  $277$ &  $43.96$ ($42.15$)  & $43.56$ ($43.28$) & $\mathbf{67.94}$ ($18.54$)  \\[0.3cm]

%
            THU-Read  \cite{tang2018multi}   &  $650$ &  $24.72$ ($41.22$)  & $33.78$ ($36.06$) & $\mathbf{48.96}$ ($21.62$)  \\[0.3cm]

            \begin{tabular}[c]{@{}c@{}}  TEgO Vanilla  \cite{lee2019hands} \\  TEgO Vanilla illu \\ TEgO Wild \\ TEgO Wild ilu  \end{tabular}  & $190$ &  \begin{tabular}[c]{@{}c@{}} $2.45$ ($77.43$) \\  $2.96$ ($72.94$) \\ $16.45$ ($47.26$) \\ $18.09$ ($57.30$) \end{tabular} &   \begin{tabular}[c]{@{}c@{}} $12.03$ ($84.36$) \\  $14.81$ ($81.02$) \\ $16.50$ ($74.22$) \\ $16.34$ ($79.55$)  \end{tabular} & \begin{tabular}[c]{@{}c@{}} $\mathbf{46.84}$ ($13.63$) \\  $\mathbf{48.30}$ ($9.20$) \\ $\mathbf{37.25}$ ($10.23$) \\ $\mathbf{54.76}$ ($8.28$)  \end{tabular} \\[0.5cm] 
            
    \midrule
            
                        Overall   &  -- &  \begin{tabular}[c]{@{}c@{}} IoU $24.55\pm15.51$  \\  Miss Rate  $38.28\pm23.46$ \end{tabular}  &  \begin{tabular}[c]{@{}c@{}} IoU $31.35\pm17.13$   \\  Miss Rate  $50.32\pm25.07$  \end{tabular} & \begin{tabular}[c]{@{}c@{}} IoU $49.97\pm14.71$ \\  Miss Rate  $14.45\pm8.21$ \end{tabular}\\[0.3cm]


  \bottomrule
\end{tabular}
\end{table*}

\subsection{Performance Metric}

Empirical results are given in terms of Jaccard Index, also known as Intersection over Union (IoU), defined as:


\begin{equation}
    IoU=\sum_{i=0}^{k}\frac{TP}{TP+FP+FN}
\end{equation}

\noindent where $k$ is the number of classes (in our case $k=2$: arms and background). IoU is computed per class and measures the ratio between intersection of two segmentation masks (the groundtruth over the predicted) over their union. Due to the great imbalance of pixels belonging to arm and background per image, we will report exclusively IoU pertaining to the arm class, in the range $0$-$100\%$. As grountruth of the available datasets do not relate entirely to whole arms, but only to hands or skin color, reported IoU could be underestimated, so we also reported $MissRate=\frac{FN}{FN+TP}$, also in the range $0$-$100\%$.

\section{Results}
\label{results}

\begin{table}[tb]
\scriptsize
\tiny
\centering
  \caption{Comparison results using EgoGesture dataset, composed of RGB and depth egocentric images in $6$ different scenes (unlike results reported in Table \ref{tab:results}). Bold indicates best IoU.}
  \label{tab:results_egogesture}
  \begin{tabular}{cccccc}
    \toprule
    Scene &  Color  & DeepLabv3+ & \begin{tabular}[c]{@{}c@{}}  DeepLabv3+ \\  EgoArm \end{tabular}  &  Depth    \\
    \hline
      \begin{tabular}[c]{@{}c@{}}  Scene1: Indoors \\  Clutter Background  \end{tabular}  &  $40.37$ ($31.86$)  & $44.64$ ($39.72$) &  $49.26$ ($23.73$)  &  $\mathbf{77.57}$ ($14.42$) \\[0.3cm] 
      \begin{tabular}[c]{@{}c@{}}  Scene2: Indoors \\  Dynamic Background  \end{tabular}  &  $48.72$ ($28.95$)  & $38.39$ ($50.78$) &  $70.33$ ($21.37$ &  $\mathbf{75.00}$ ($20.01$) \\[0.3cm] 
            \begin{tabular}[c]{@{}c@{}}  Scene3: Indoors \\  Toward Windows  \end{tabular}  &  $21.40$ ($70.54$)  & $25.48$  ($71.75$) &  $67.32$ ($21.54$) &  $\mathbf{73.91}$  ($20.64$) \\[0.3cm] 
 \begin{tabular}[c]{@{}c@{}}  Scene4: Indoors \\  Walking  \end{tabular}  &  $47.25$ ($37.13$ )& $35.98$ ($37.47$)   & $62.93$ ($20.05$) & $\mathbf{76.54}$ ($19.25$) \\[0.3cm] 
  \begin{tabular}[c]{@{}c@{}}  Scene5: Outdoors \\  Dynamic Background  \end{tabular}  &  $49.54$ ($46.79$) & $57.50$ ($28.53$) &   $\mathbf{77.09}$ ($10.76$)  &  $72.57$ ($24.22$) \\[0.3cm] 
    \begin{tabular}[c]{@{}c@{}}  Scene6: Outdoors \\  Walking \\ Dynamic Background  \end{tabular}  &  $56.69$ ($38.78$) &  $61.58$ ($29.83$) & $\mathbf{79.97}$ ($12.22$) &  $69.79$ ($28.21$) \\[0.3cm]     
  \bottomrule
\end{tabular}
\end{table}

Table \ref{tab:results} reports the Sem-Seg results in terms of IoU and Miss Rate for the arm class, using three different segmentation algorithms based on color or deep learning and for the test datasets reported in Section \ref{test_databases}. Also, Table~\ref{tab:results_egogesture} further indicates IoU reached on each of the $6$ scenes of EgoGesture. 

\subsection{Color Performance}

Color-based segmentation is applied using an HSV filtering similar to Equation \ref{green_chroma}, where Hue values are around the skin color. As can be seen from Table \ref{tab:results}, color-based segmentation achieves similar or worse results than the original DeepLabv3+. Concretely, there is a range of absolute improvement from $10.00\%$ to $25.00\%$ when replacing color-based to DeepLabv3+ for GTEA Gaze+, EDSH, THU-Read and TeGO datasets. This is logically expected, as this approach considers exclusively the color for making its decision. Performance is also hindered when there are objects in the scene which share the skin color; notice the very bad performance achieved with the GTEA Gaze+ or THU-Read databases due to their yellowish / reddish scene appearance (see Fig.\ref{fig:all_datasets} A and J, respectively). Also, results reported from the TEgO database show that relying exclusively on color is not an appropriate method for applications where there are people from different ethnicities, (see Fig.\ref{fig:samples_tego}). For the case of EgoGesture reported in Table \ref{tab:results}, there is not observable differences between the color performance and the original DeepLabv3+. However, when assessing those results per scene (see Table \ref{tab:results_egogesture}), we observe: both color and DeepLabv3+ are severely affected by extreme illumination conditions (Scene3, see Fig.\ref{fig:egogesture}B); color-based is more robust than DeepLabv3+ in both dynamic or walking indoor scenarios where movement can produce some blur effect (around $10.00\%$ average absolute improvement when using color-based rather than DeepLabv3+ for Scene2 and Scene4, see Fig.\ref{fig:egogesture}A), and that DeepLabv3+ outperform color when good illumination is available (Scene5 and Scene6) or there is controlled background.

\subsection{Deep performance}

Concerning the behavior of the two different deep segmentation networks assessed, we observe the general superiority of the networks using EgoArm in comparison with the original DeepLabv3+ network. This observation validates our hypothesis of the convenience of adapting the network with a database more similar to the real application.

We observe slight, moderate or large improvement, depending on the particular dataset. In a high level perspective, this is expected since deep learning algorithms, unlike color-based segmentation, are not only considering color information for the segmentation task, but also complementary information such as shapes, texture or more abstract and complex information.

\begin{figure*}[h]
  \centering
  \includegraphics[width=1.0\linewidth]{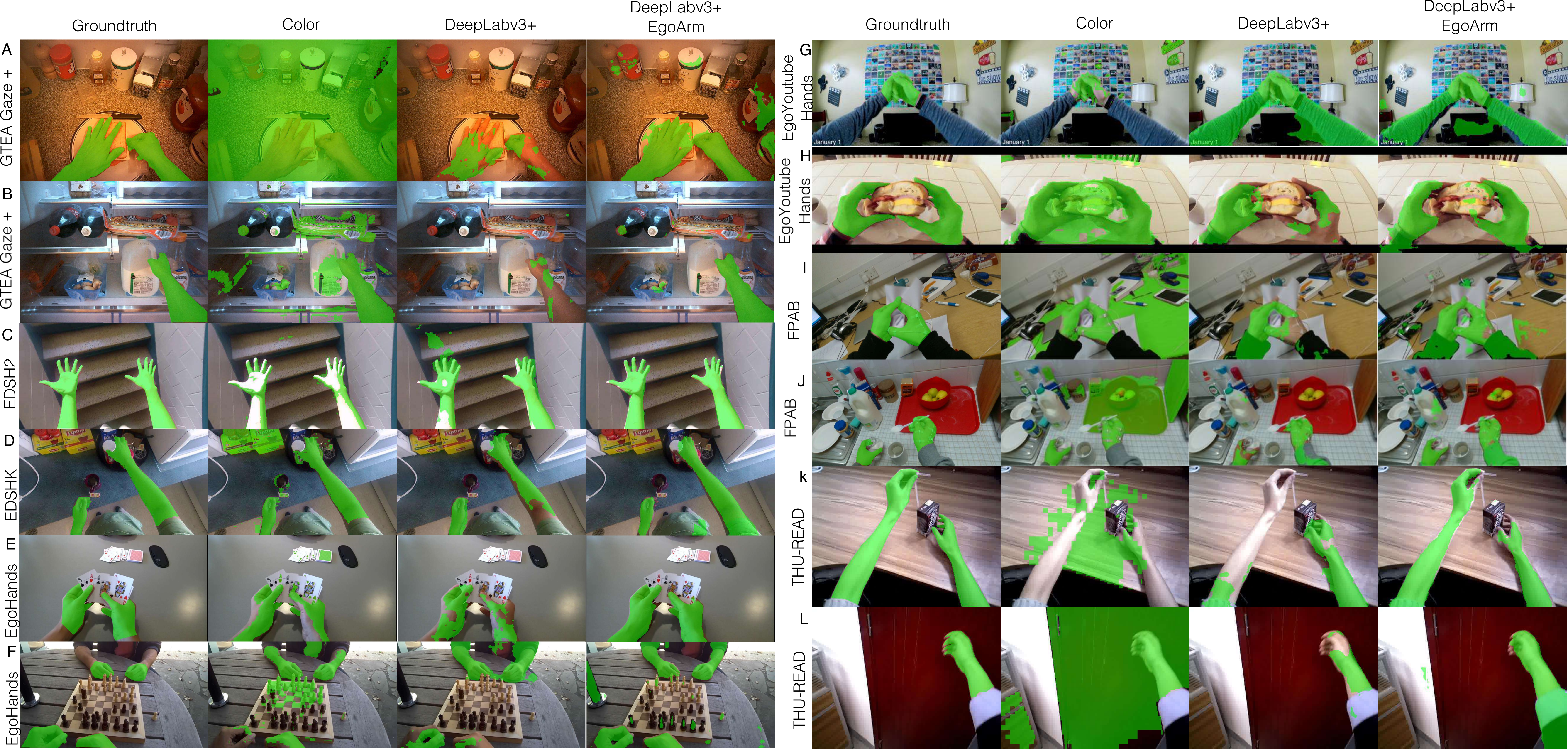}
  %
  %
  \caption{\label{fig:all_datasets}
    Segmentation examples of the different segmentation methods explored for GTEA GAZE+, EDSH, EgoHands, EgoYoutubeHands,TeGO, FPAB and THU-Read. The first column refers to the groundtruth defined in each case. In all cases, groundtruth are related to just hands or skin color, whereas our target is detecting the whole arm. Notice that FPAB do not contain groundtruth images, we have manually labeled these two examples.}
\end{figure*}

Slight improvement is observed for EgoHands ($6.87\%$ absolute improvement) and no improvement is observed for the Ego Youtube Hand Datasets. In both cases, the overall segmentation results from these two datasets are very poor due to the groundtruth being related just to hands despite the majority of images contain whole arms with or without clothes (see Fig.\ref{fig:all_datasets} G and F). Also, in the case of EgoHands (see Fig.\ref{fig:all_datasets} F), the majority of images present both egocentric and third POV arms. In general, third POV arms occupy a larger surface.~As the networks trained with EgoArm are focused on egocentric arms, it is logically that there is not a huge improvement with respect to the original DeepLabv3+.~In what concerns EgoYoutubeHands images, we assume that the low resolution of these images ($384\times216$) along with the very uncontrolled and cluttered environment makes the segmentation very challenging.

Concerning results reported on EDSH2, there is also a slight gain when including EgoArm ($6.50\%$ absolute improvement). We believe this is because most of these test images just contain arms but not clothes, and also such images are controlled both in terms of the background but also regarding the hands (e.g. fingers are normally very well separated). For the more uncontrolled case of EDSHK, there is a larger improvement specially in terms of MissRate (from $22.91$ to $8.10$) between the original DeepLabv3+ and the one trained with EgoArm. Specially in the egocentric images from EDSHK, there is very frequent to encounter arms with clothes, which are not considered part of the groundtruth (see Fig.\ref{fig:all_datasets} D). Therefore, part of the false positive rate is related to the clothe side of the arm. 

Moderate enhancement is encountered for the GTEA Gaze+ validation dataset, THU-Read, and EgoGesture (in the range of $15.00\%$ to $25.00\%$ absolute improvement). As these datasets are purely egocentric, it is more noticeable the gain achieved when including the EgoArm (see Fig.\ref{fig:all_datasets} A-B, Fig.\ref{fig:all_datasets} K-L or Fig.\ref{fig:egogesture},).  Also, in the case of EgoGesture, the fact that egocentric scenes are very clean with not objects surrounding the arms prevents additional mistakes. Having a more in-depth look to the IoU reported per scene in Table \ref{tab:results_egogesture}, it is also observed the gain achieved with the deep network trained on EgoArm in all scenes and notably in outdoors scenarios, which, apart from the aforementioned nature of the scenarios, is probably due to the uniform and good illumination. Lastly, a huge increase of performance is observed with the different subsets from TEgO dataset (in the range of $20-40\%$ of absolute improvement). The main reason behind it relies on the diversity of skin colors presented in the EgoArm.

\begin{figure}[t]
  \centering
  \includegraphics[width=.9\linewidth]{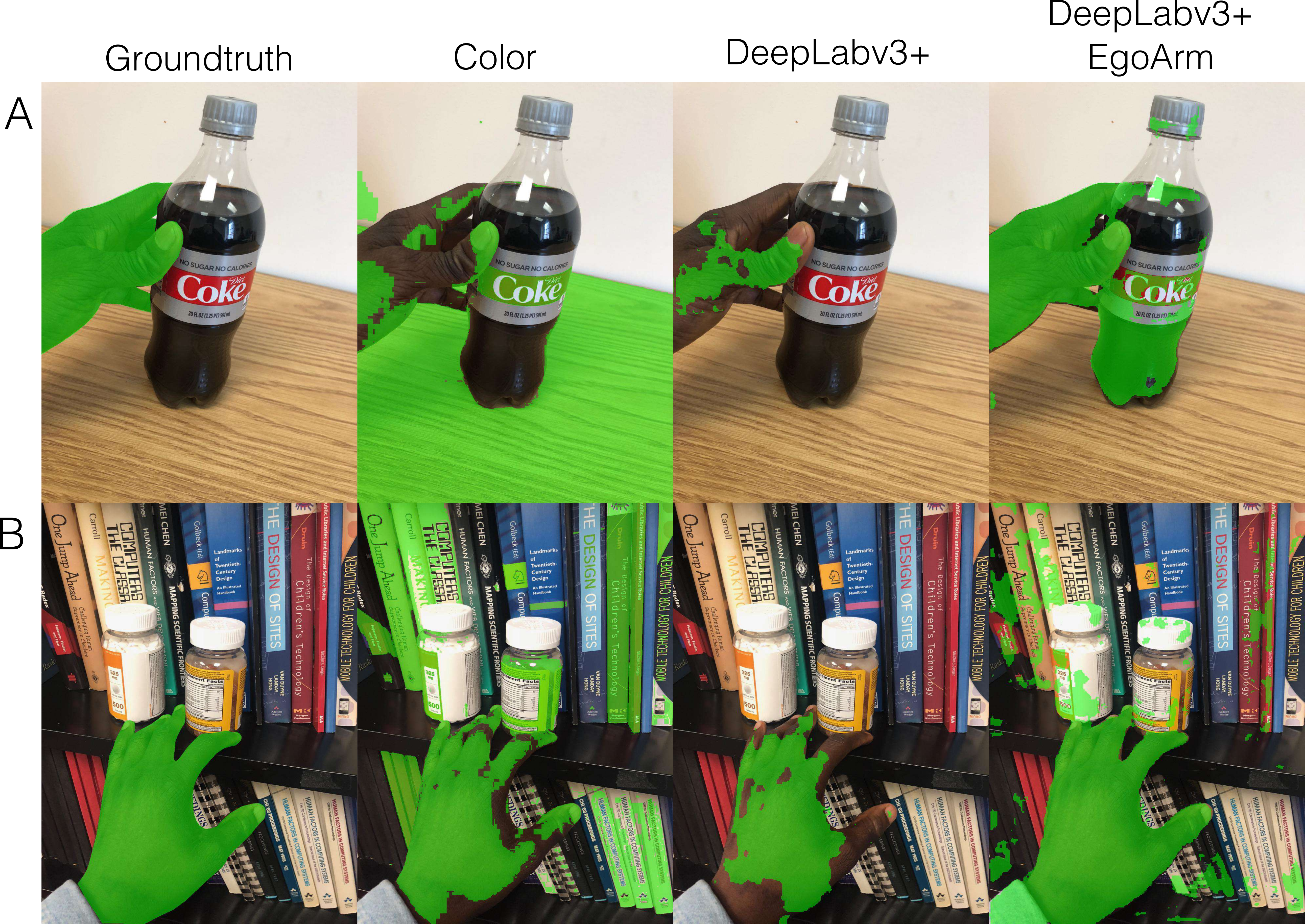}
  %
  %
  \caption{\label{fig:samples_tego}
           Samples from TeGO. Notice that the groundtruth is related to the skin color, so clothes are not considered.}
\end{figure}

After having a visual inspection to the images, we notice that in some cases, the DeepLabv3+ EgoArm network generates some false positives from background items with some color similarities. We deduce that the nature of semi-synthetic images of EgoArm, combining arms with a wide variety of natural backgrounds, is not always fully representing real nature and coherent scenes, preventing the encoding subnetwork (focused on pixel classification) to optimize its performance, generating some false positives that are later upsampled through the decoding subnetwork and the short-chut connections (see Fig.\ref{fig:samples_tego} A where part of the coke is also segmented as arms or red color items from the kitchen scenes presented in Fig.\ref{fig:all_datasets}).  We believe this false positive problem can be overcome by improving the classification performance of the encoding subnetwork \cite{li2019supervised} and by a more in-depth assessment of which background type are more appropriate for targeting real AV applications.


\subsection{Comparison with depth}
\label{compare_with_depth}

As stated in Section \ref{related_works}, segmentation based on depth implies select all objects from the user surroundings that are below a particular distance threshold. Here we aim to compare the results obtained with depth in comparison with color-based or deep-based segmentation, using the subset of EgoGesture described in Section \ref{test_databases}. 



\begin{figure}[t]
  \centering
  \includegraphics[width=1.0\linewidth]{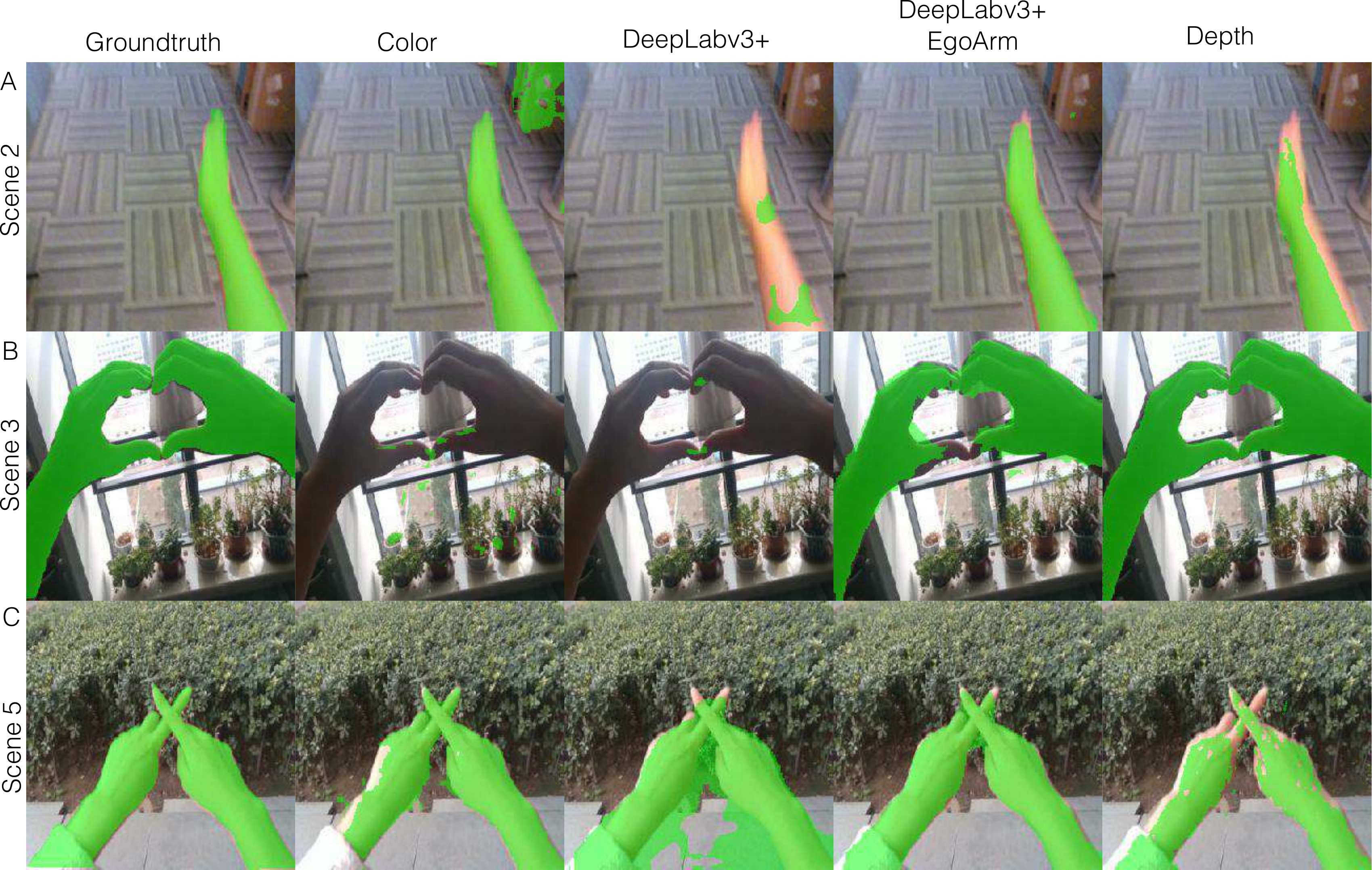}
  %
  %
  \caption{\label{fig:egogesture}
           Comparison results using EgoGesture dataset, composed of RGB and depth egocentric images in $6$ different scenes. Here the groundtruth refers to the whole arm (we manually labelled them).}
\end{figure}

It is clearly visible from Table \ref{tab:results_egogesture} that segmentation based on depth is more uniform across the different scenes than deep or color-based approaches. A slight drop in depth performance is shown in outdoors scenarios (see for instance Fig.\ref{fig:egogesture} C), possibly because signal light (texture being projected in infrared to compute depth through disparity) is much weaker than ambient sunlight. However, despite this general superior performance, we believe EgoGesture images do not represent real and challenging scenarios concerning depth estimation. EgoGesture depth estimation works fine because the dataset generation scenes avoid all the critical scenarios for RGB-D sensors \cite{moya2017modeling}: hands are always within the distance range of the camera (and never closer) and no other object is in such range, hands are fully visible from both infrared sensors, and they never cast shadows from the infrared emitter. Moreover, there are recent studies exploring deep learning to enhance depth maps (also known as depth completion), that suggest that there is still a large room for improvement in this area \cite{zhang2018deep,ma2019self,li2018undeepvo}. Once achieved, it could reliably segment near objects in AV applications. 

\subsection{Computation time}

 
Given a $720\times720$ image, segmenting it with color, depth, and deep approaches would take $2.9ms$, $700\mu s$\footnote{This does not include the time required to generate the depth map from the stereoscopic images.} and $74ms$, respectively using a PC Intel Xeon ES-2620 V4 @ $2.1$Ghz with $32$ GB powered with $2$ GPU GTX-1080 Ti with $12GB$ RAM. Our deep implementation achieves about $15$ fps, which is 4 to 6 times slower than what it would be desirable for a smooth AV system.
However, it is within the right order of magnitude, and therefore it is just a matter of algorithm optimization and hardware improvement 
that the Sem-Seg approach can work in real time. In practice, it would imply either having the HMD attached to a resourceful computer or offloading computation to the edge cloud \cite{li2018edge,erol2018caching}.

\section{Conclusion}
\label{conclusions}


In this study, we have proposed the use of deep learning to segment egocentric human body parts, in particular arms, to enhance self-perception in AV. We have conducted first a thorough survey based on existing egocentric segmentation methods, mainly based on color or depth. In order to target the requirements of first POV segmentation for AV, we have created the \textit{EgoArm: Egocentric Arm Segmentation dataset} composed of more than $10,000$ images including variations of gender, arm positions, clothes, indoors and outdoors, and skin color along with a procedure to generate automatic groundtruth. Later, we have explored different semantic segmentation networks to target egocentric arm segmentation. We have reported results on different real egocentric datasets: GTEA Gaze+, EDSH, EgoYoutubeHands, EgoHands, FPAB, THU-Read, TEgO, and Ego Gesture, providing comparisons with color- and depth-based segmentations. Results have proven the effectiveness of EgoArm for arm segmentation, boosting the average IoU from $25.00\%$ IoU reached with chroma or from $31.35\%$ reached with the original DeepLabv3+ network, up to $50.00\%$ IoU. Besides, these segmentation networks are more robust than color-based segmentation at dealing with illumination changes, segmeting clothes or arm with different skin color, etc. In comparison with depth, deep-based segmentation algorithms are able to segment the desired objects exclusively, while depth will be forced to segment everything below a distance-threshold, which is a paradigm that may not apply to all AV applications. Besides, despite not being shown in the EgoGesture database, there are current challenges at estimating depth, that would need to be solved before using it reliably for AV. 

\ifCLASSOPTIONcaptionsoff
  \newpage
\fi



%

\bibliography{egbibsample}
\bibliographystyle{abbrv-doi}

%
%

%

\begin{IEEEbiography}[{\includegraphics[width=1in,height=1.25in,clip,keepaspectratio]{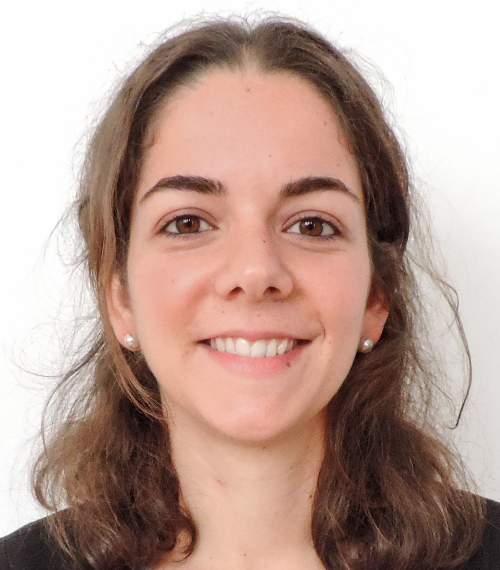}}]%
{Ester Gonzalez-Sosa}
received the B.S. in computer science and M.Sc in Electrical Engineering from Universidad de Las Palmas de Gran Canaria in $2012$ and $2014$, respectively. In June $2017$ she obtained her PhD degree from Universidad Autonoma de Madrid, within the Biometric Recognition Group. In October 2017 she joined the Distributed Reality Solutions Lab at Nokia Bell-Labs. She has carried out several research internships in worldwide leading groups in biometric recognition such as TNO, EURECOM, or Rutgers University. Her research interests include biometrics with emphasis on face, body, soft biometrics and millimeter imaging, and computer vision techniques applied to egocentric perception and Mixed Reality applications. Gonzalez-Sosa has been the recipient of the competitive Obra Social La CAIXA Scholarship ($2012$), the UNITECO AWARD from the Spanish Association of Electrical Engineers ($2013$) and the European Biometrics Research Award ($2018$).
\end{IEEEbiography}

\begin{IEEEbiography}[{\includegraphics[width=1in,height=1.25in,clip,keepaspectratio]{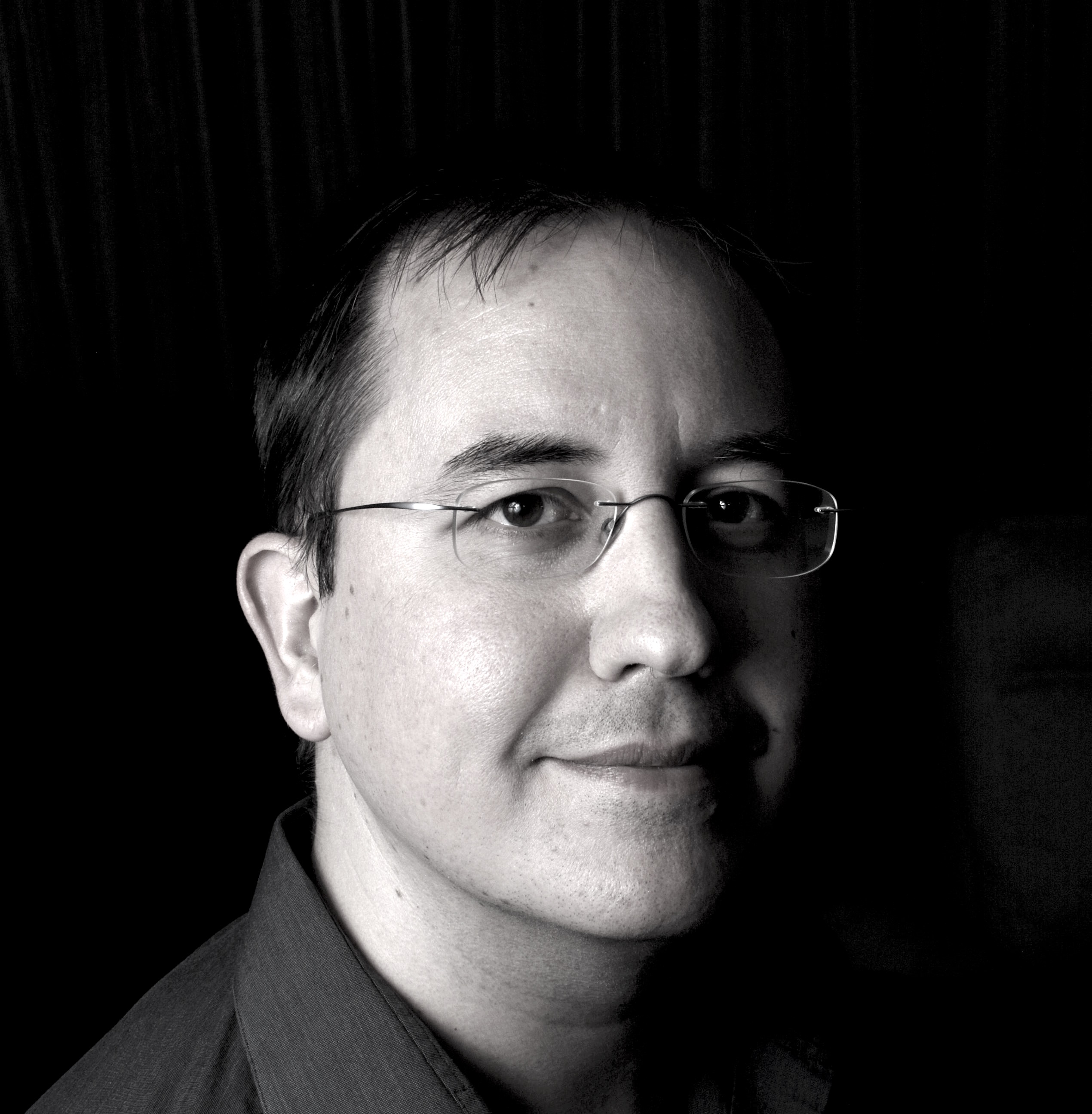}}]
{Pablo Perez}
received the Telecommunication Engineering degree (integrated BSc-MS) in 2004 and the Ph.D. degree in Telecommunication Engineering in 2013 (Doctoral Graduation Award), both from Universidad Politécnica de Madrid (UPM), Madrid, Spain. From 2004 to 2006 he was a Research Engineer in the Digital Platforms Television in Telefonica I+D and, from 2006 to 2017, he has worked in the R\&D department of the video business unit in Alcatel-Lucent (later acquired by Nokia), serving as technical lead of several video delivery products. Since 2017, he is Senior Researcher in the Distributed Reality Solutions department at Nokia Bell Labs. His research interests include multimedia quality of experience, video transport networks, and immersive communication systems.
\end{IEEEbiography}

\begin{IEEEbiography}[{\includegraphics[width=1in,height=1.25in,clip,keepaspectratio]{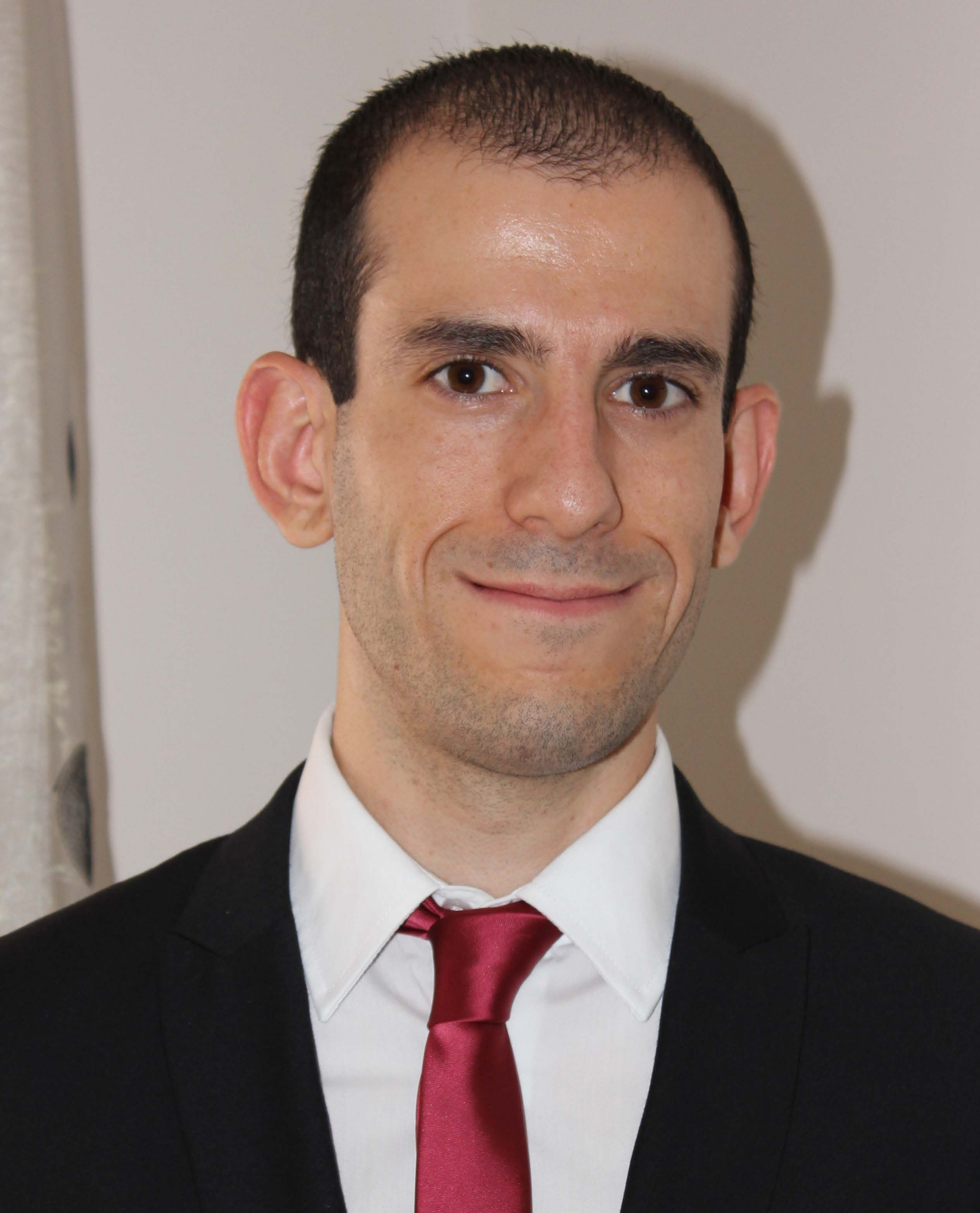}}]%
{Ruben Tolosana}
received the M.Sc. degree in Telecommunication Engineering, and his Ph.D. degree in Computer and Telecommunication 
Engineering, from Universidad Autonoma de Madrid, in 2014 and 2019, respectively. In April 2014, he joined the Biometrics and Data Pattern Analytics - BiDA Lab at the Universidad Autonoma de Madrid, where he is currently collaborating as a Postdoctoral researcher. Since then, Ruben has been granted with several awards such as the FPU research fellowship from Spanish MECD (2015), and the European Biometrics Industry Award (2018). His research interests are mainly focused on signal and image processing, pattern recognition, and machine learning, particularly in the areas of face manipulation, human-computer interaction, and biometrics. He is author of several 
publications and also collaborates as a reviewer in many different high-impact conferences (e.g., ICDAR, IJCB, ICB, BTAS, EUSIPCO, etc.) 
and journals (e.g., IEEE TPAMI, TCYB, TIFS, TIP, ACM CSUR, etc.). Finally, he has participated in several National and European projects 
focused on the deployment of biometric security through the world
\end{IEEEbiography}

\begin{IEEEbiography}[{\includegraphics[width=1in,height=1.25in,clip,keepaspectratio]{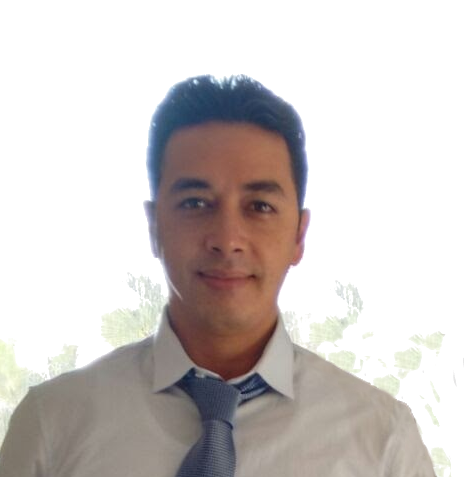}}]%
{Redouane Kachach}
Redouane Kachach is a researcher in Nokia Bell Labs department for Distributed Reality Solutions, working mainly on the next generation of immersive video technologies. As professional background, Redouane is a senior software engineer with a large experience in software design and implementation of real-time, distributed, large-scale video processing and distribution systems. He joined joined Nokia Bell-Labs in 2017 as research engineer. Currently, he is working on the next generation of human communication solutions based mainly on immersive video and mixed reality.
\end{IEEEbiography}

\begin{IEEEbiography}[{\includegraphics[width=1in,height=1.25in,clip,keepaspectratio]{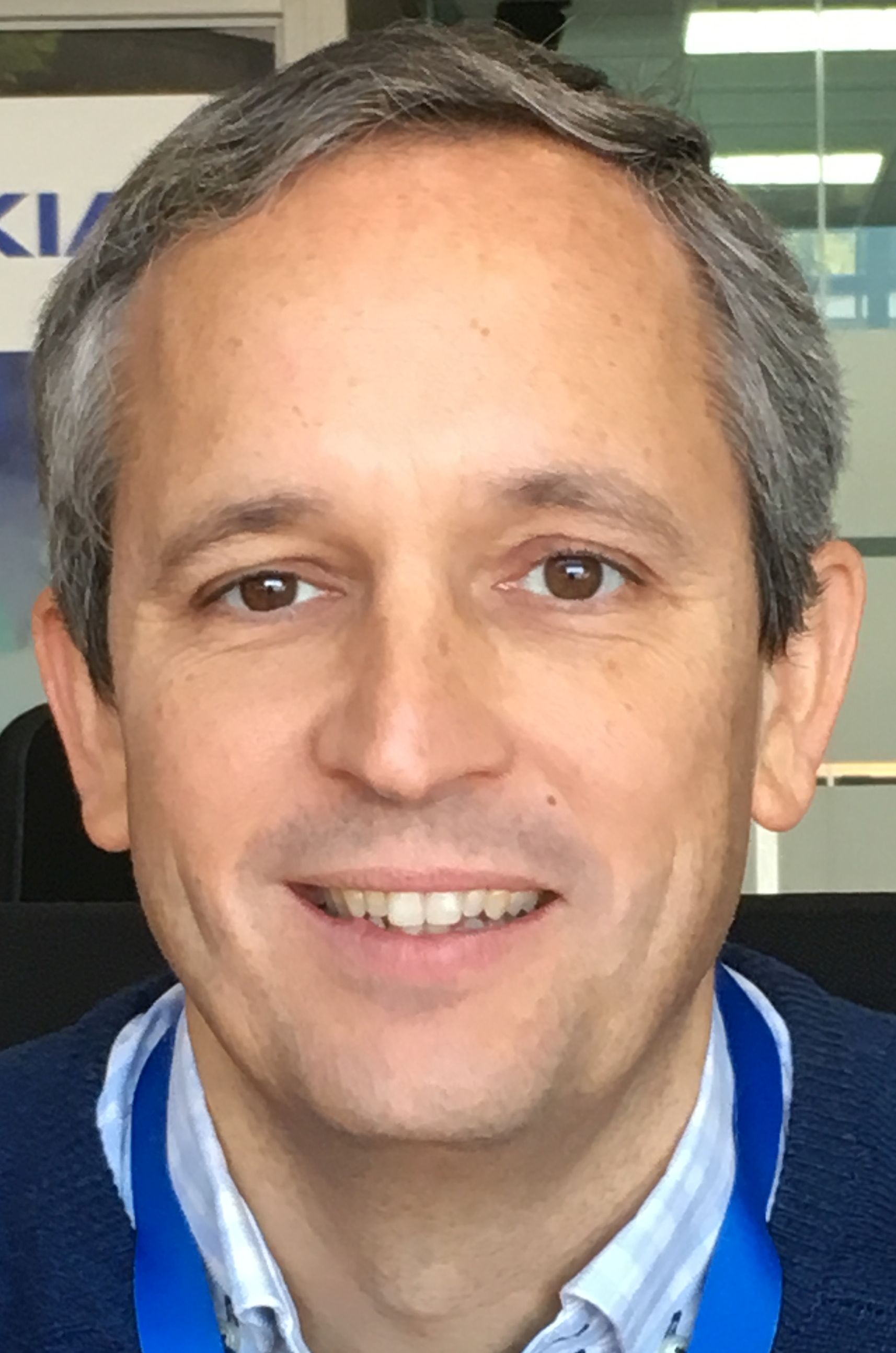}}]%
{Alvaro Villegas} is the Head of Bell Labs in Nokia Spain, a research center focused on the application of immersive media to human communications. He received a six-year telecommunications engineering degree at Universidad Politécnica de Madrid (Spain), and he completed an MBA Core Program at ESCP Europe Business School. For the last twelve years he has worked as an innovation lead in the field of digital video first in Lucent, then in Alcatel-Lucent and now in Nokia, where he has filed more than $40$ patents in the field. He was awarded by Bell Labs with the Distinguished Member of Technical Staff title. His prior professional experience, always in the field of video innovation, was developed in Siemens, Telefonica R\&D, ONO, Motorola and Nagravision.
\end{IEEEbiography}








\end{document}